\documentclass[11pt,a4paper]{article}
\usepackage[hyperref]{acl2020}
\usepackage{times}
\usepackage{graphicx}
\usepackage{dirtytalk}
\usepackage{adjustbox}
\usepackage{float}
\usepackage{latexsym}
\usepackage{tikz}
\usepackage{amsmath}

\usetikzlibrary{fit, shapes, arrows.meta, positioning}

\makeatletter
\newcommand{\printfnsymbol}[1]{%
  \textsuperscript{\@fnsymbol{#1}}%
}
\makeatother

\usepackage{microtype}

\aclfinalcopy % Uncomment this line for the final submission
\def\aclpaperid{127} %  Enter the acl Paper ID here

\title{How much complexity does an RNN architecture need to learn syntax-sensitive dependencies?}

\author{
Gantavya Bhatt\thanks{$^*$Equal Contribution} \textsuperscript{ 1}, Hritik Bansal\printfnsymbol{1}\textsuperscript{1}, Rishubh Singh\printfnsymbol{1}\textsuperscript{1}%\thanks{Work done as a student at IIT Delhi.}
, Sumeet Agarwal\textsuperscript{1}\\
  \textsuperscript{1} Indian Institute of Technology Delhi\\
  \texttt{\{gantavya.iitd, hbansal10n, rishubhsingh135\}@gmail.com} \\
  \texttt{sumeet@iitd.ac.in} 
}
\date{}

\begin{document}
\maketitle
\begin{abstract}
Long short-term memory (LSTM) networks and their variants are capable of encapsulating long-range dependencies, which is evident from their performance on a variety of linguistic tasks. On the other hand, simple recurrent networks (SRNs), which appear more biologically grounded in terms of synaptic connections, have generally been less successful at capturing long-range dependencies as well as the loci of grammatical errors in an unsupervised setting. In this paper, we seek to develop models that bridge the gap between biological plausibility and linguistic competence. We propose a new architecture, the {\em Decay RNN}, which incorporates the decaying nature of neuronal activations and models the excitatory and inhibitory connections in a population of neurons. Besides its biological inspiration, our model also shows competitive performance relative to LSTMs on subject-verb agreement, sentence grammaticality, and language modeling tasks. These results provide some pointers towards probing the nature of the inductive biases required for RNN architectures to model linguistic phenomena successfully.
%Long short-term memory (LSTM) and its variants, a group of recurrent networks, are capable of encapsulating long term dependencies, which is evident by their performance on the linguistic tasks. On the other hand, a simple recurrent network (SRN), which has more biological grounding in terms of synaptic connections, has consistently failed on capturing long term dependencies as well as the locus of grammatical errors in an unsupervised setting. In this paper, we use insights from psycholinguistics to develop a model that bridges the gap between neurological persuasiveness and commendable performance on linguistic tasks. We propose a new architecture- \say{Decay RNN} which incorporates the decaying nature of voltage in a neuron membrane at the same time modeling the excitatory and inhibitory connections in the population of neurons. Besides its biological plausibility, our model also has a competitive performance to LSTM (and its variants) on subject-verb agreement, sentence grammaticality, and language modeling tasks. The results on our model open the room to probe the nature of inductive biases responsible for the laudable or substandard performance of certain architectures.
\end{abstract}

\section{Introduction}
\label{sec:Introduction}
% Early notable works on the neural networks (NN) such as Hebbian Learning \cite{hebb1949organization} and the Perceptron \cite{rosenblatt1958perceptron} described biologically inspired learning of parameters. Another wave of Connectionism came with the invention of backpropagation, which is still the backbone of neural network training. 
For the last couple of decades, neural networks have been approached primarily from an engineering perspective, with the key motivation being efficiency, consequently moving further away from biological plausibility. Recent developments \cite{song2016training, gao2015simplicity,sussillo2013opening} have however incorporated explicit constraints in neural networks to model specific parts of the brain and have found a correlation between the learned activation maps and actual neural activity recordings. Thus, these trained networks can perhaps act as a proxy for a theoretical investigation into biological circuits.

%Chomsky's theory of innate language structures in the human mind \cite{chomsky1965, Chomsky1980-CHORAR-5} and the claims on hierarchical bias in children's language acquisition have influenced the development of psycholinguistics and the way it aids in developing the representation of human cognition. 
Recurrent Neural Networks (RNNs) have been used to analyze the principles and dynamics of neural population responses by performing the same tasks as animals \cite{Mante2013}. However, these networks violate Dale's law \cite{dale1935pharmacology, strata1999dale}, which states that the neurons have either a purely excitatory or inhibitory effect on other neurons in the mammalian brain. The decaying nature of the potential in the neuron membrane after receiving signals (excitatory or inhibitory) from the surrounding neurons is also well-studied \cite{gluss1967model}. The goal of our work is to incorporate these biological features into the RNN structure, which gives rise to a neuro-inspired and computationally inexpensive recurrent network for language modeling, which we call a {\em Decay RNN} (Section \ref{sec:Decay RNN}). We perform learning using the backpropagation algorithm. Despite its differences with the way learning is believed to happen in the brain, it has been argued that the brain can implement its core principles \cite{hinton2007backpropagation, Lillicrap2020}.
% \textcolor{blue}{The goal of our work is to introduce a neuro-inspired and computationally inexpensive recurrent network for language modeling, which we call a {\em Decay RNN} (Section \ref{sec:Decay RNN})}. 
We assess our model's ability to capture syntax-sensitive dependencies via multiple linguistic tasks (Section \ref{sec:Experiments}): number prediction, grammaticality judgement \cite{linzen2016assessing} which entails subject-verb agreement, and a more complex language modeling task \cite{marvin2018targeted}.

Subject-verb agreement, where the \textit{main noun} and the \textit{associated verb} must agree in number, is considered as evidence of hierarchical structure in English. This is exemplified using a sentence taken from the dataset made available by
%Subject-verb agreement, where the \textit{main noun} and the \textit{associated verb} must agree in number, is considered as an evidence of hierarchical structure in English. natural language. This is exemplified using a sentence taken from the dataset made available by
\citet{linzen2016assessing}:

\begin{enumerate}
    \item *All \textbf{trips} on the \underline{expressway} \textbf{requires} a toll.
    \item All \textbf{trips} on the \underline{expressway} \textbf{require} a toll.
\end{enumerate}

The effect of agreement attractors (nouns having number opposite to the main noun; \textit{expressway} in the above example\footnote{Main noun and verb are highlighted in bold. Intervening nouns are underlined. Asterisks mark unacceptable sentences.}) between the main noun and main verb of a sentence has been well-studied \cite{linzen2016assessing,kuncoro-etal-2018-lstms}. Our work also highlights the influence of non-attractor intervening nouns. For example,
\begin{itemize}
    \item A \textbf{chair} created by a \underline{hobbyist} as a \underline{gift} to \underline{someone} \textbf{is} not a commodity.\footnote{\label{note2}Sentence taken from the dataset made available by \citet{linzen2016assessing}.}
\end{itemize}
% We assess our model's performance in the presence of such intervening nouns, by making it predict the number of the verb, and judge the grammaticality of a sentence in Section \ref{sec:Experiments}.
% We compare our model to both LSTMs and SRNs on the verb number prediction and grammaticality tasks.
In the number prediction task, if a model correctly predicts the grammatical number of the verb (singular in case of `is'), it might be due to the (helpful) interference of non-attractor intervening nouns (`hobbyist', `gift', `someone') rather than necessarily capturing its dependence the main noun (`chair'). From our investigation in Section \ref{subsec:Joint analysis}, we find that the linear recurrent models take cues present in the vicinity of the main verb to predict its number, apart from the agreement with the main noun.

In the subsequent sections, we investigate the performance of the Decay RNN and other recurrent networks, showing that no single sequential model generalizes well on all (grammatical) phenomena, which include subject-verb agreements, reflexive anaphora, and negative polarity items as described in \citet{marvin2018targeted}.
%For a robust examination of our proposed model, in Section \ref{subsec:Language Modeling} we present our model’s performance on word-level language modeling task and will present that even with a minimalistic change in the number of parameters from SRN, our model achieved strong performance jump on an exponential scale.
%In section \ref{subsec: Targeted Syntactic Evaluation}, we show that there is no recurrent network that generalizes well on all types of sentences, which include subject-verb agreements, reflexive anaphoras, and negative polarity items as described in \citet{marvin2018targeted}. This indicates the need for further research on the nature of the inductive bias conferred by certain classes of models and training procedures, which leads to variability in performance across sentence types.
Our major outcomes are: 

\begin{enumerate}
    \item Designing a relatively simple and bio-inspired %\textcolor{blue}{biologically plausible}
    recurrent model: the Decay RNN, which performs on-par with LSTMs for linguistic tasks such as subject-verb agreement and grammaticality judgement.   
    %\item Designing a neurologically plausible and mathematically lucid recurrent scheme-Decay RNN, which performs on par with LSTM over linguistic tasks such as subject-verb agreement, grammaticality, and language modeling.   
    \item Pointing to some limitations of analyzing the intervening attractor nouns alone for the subject-verb agreement task and attempting joint analysis of non-attractor intervening nouns and attractor nouns in the sentence.
    \item Showing that there is no linear recurrent scheme which generalizes well on a variety of sentence types and motivating research in better understanding of the nature of biases induced by varied RNN structures.
\end{enumerate}

\section{Related Work}
\label{sec:Related Work}
There has been prior work on using LSTMs \cite{hochreiter1997long} for language modeling tasks. The work of \citet{gers2001lstm} has shown that LSTMs can learn simple context-free and context-sensitive languages. However, as per the investigations carried out in \citet{kuncoro-etal-2018-lstms}, it was observed that if the model capacity is not enough, then LSTMs may not generalize the long-range dependencies. Recently many architectures have explicitly incorporated the knowledge of phrase structure trees \cite{kuncoro-etal-2018-lstms, alvarez2016tree, tai-etal-2015-improved} which have shown improvement in generalizing over long-range dependencies. At the same time, \citet{shen2018ordered} proposed ON-LSTMs, a modification to LSTMs that provides an inductive tree bias to the structure. However, \citet{dyer2019critical} have shown that the success of ON-LSTMs was due to their proposed metric to analyze the model, not necessarily due to their architecture. 

From the biological point of view, \citet{capano2015optimal} used a hard reset of the membrane potential in contrast to a soft decay observed in a neuronal membrane. At the same time, their learning paradigm is similar to the Hebbian learning scheme \cite{hebb1949organization}, which does not involve error backpropagation \cite{rumelhart1986learning}.
Our work is closely related to the idea of modeling the population of neurons as a dynamical system (EIRNN) proposed by \citet{song2016training}. However, their time constant parameter was based on the concepts described in \citet{wang2002probabilistic} while the sampling rate was arbitrarily chosen. Given that the chosen values only considered a certain class of neurons \cite{yang2019task}, we believe that it is not necessary to have the same values of the parameters for each cognitive task. Thus, we build on their formulation by making the sampling rate and time constant learnable as manifested by our decay parameter, described in the next section. 
% \textcolor{red}{We also address whether the models proposed for low-level tasks, as described in \citet{song2016training}, perform well on the high-level cognitive tasks outlined in the psycholinguistic literature.} 

% CAN REMOVE THE LINE BELOW?
%Many of the models proposed so far are more complex than LSTMs and are far away from being cognitively plausible. Hence, we aim to address the issue of achieving LSTM-like performance with a simple and biologically plausible model in our paper. 
\section{Biological Preliminaries}

According to Dale's principle, a neuron is either excitatory or inhibitory \cite{eccles1976electrical}. If a neuron output produces a negative (positive) change in the membrane potential of all the connected neurons via its synapse, then it is said to be an inhibitory (excitatory) neuron. In a set of $N$ neurons, if $\mathbf{W}$ is the synaptic connection matrix, then the connection from the neuron $j$ to neuron $i$ is `excitatory' if $W_{ij} > 0$, and `inhibitory' if $W_{ij} \leq 0$.  \citet{capano2015optimal} have argued that a balance between structural and response variability (entropy), and excitability (synaptic strength) of a network maximizes the overall learning. This balance is governed by the ratio of inhibitory and excitatory neurons. They have further shown that this balance also maximizes the overall performance in multitask learning. \citet{catsigeras2013dale} mathematically prove that Dale's principle is necessary for an optimal\footnote{In the sense of showing the most diverse set of responses.} neuronal network's dynamics.
 
In the postsynaptic neuron, the integration of synaptic potentials is realized by the addition of excitatory (+ve) and inhibitory (-ve) postsynaptic potentials (PSPs). PSPs are electronic voltages, that decay as a function of time due to spontaneous reclosure of the synaptic channels. The decay of the PSPs is controlled by the membrane constant $\tau$, i.e., the time required by the PSP to decay to 37\% of its peak value \cite{Wallisch2009}.

\section{Decay RNN}
\label{sec:Decay RNN}
Here we present our proposed architecture, which we call the {\em Decay RNN} (DRNN). Our architecture aims to model the decaying nature of the voltage in a neuron membrane after receiving impulses from the surrounding neurons. At the same time, we incorporate Dale's principle in our architecture. Thus, our model captures both the microscopic and macroscopic properties of a group of neurons. Adhering to the stated phenomena, we define our model with the following update equations for given input \(\mathbf{x}^{(t)}\) at time $t$:
% \textcolor{blue}{Here we present our proposed architecture which we call the \say{Decay RNN} (DRNN). Through our architecture, we aim to encapsulate the decaying nature of the voltage in a neuron membrane after receiving excitatory and inhibitory impulses from the surrounding neurons \cite{gluss1967model}. At the same time, we also incorporate Dale's principle \cite{dale1935pharmacology} which is thought to be a fundamental property of populations of neurons. According to Dale's principle, neurons exhibit either purely excitatory or inhibitory effects over postsynaptic neurons. Empirically, the excitatory effect outnumbers the inhibitory with a ratio close to 4:1 \cite{capano2015optimal}. Adhering to the above phenomena, we came up with the following update equations for given input \(\textbf{x}^{(t)}\) at time $t$: }
\begin{center}
    $\mathbf{c}^{(t)}=
    (ReLU(\mathbf{W})\mathbf{W}_{dale})\mathbf{h}^{(t-1)}
    +\mathbf{U}\mathbf{x}^{(t)} + \mathbf{b}$
    
    $\mathbf{h}^{(t)}=f(\alpha\mathbf{h}^{(t-1)} + (1-\alpha)\mathbf{c}^{(t)})$
\end{center}

Here $f$ is a nonlinear activation function, $\mathbf{W}$ and $\mathbf{U}$ are weight matrices, $\mathbf{b}$ is the bias and \(\mathbf{h}^{(t)}\) represents the hidden state (analogous to voltage). We define $\alpha \in$ (0,1) as a learnable parameter to incorporate a decay effect in the hidden state (analogous to the decay in the membrane potential). Here $\alpha$ acts as a balancing factor between the hidden state \(\mathbf{h}^{(t-1)}\) and \(\mathbf{c}^{(t)}\).\footnote{It was kept bounded using a sigmoid function. Our results did not change when we used a linear function instead.}
% \textcolor{red}{Empirically, the fraction of inhibitory connections in the mammalian brain is suggested to be 20-30\%.}
\(\mathbf{W}_{dale}\) is a diagonal matrix, and based on the empirical results on the mammalian brain \cite{hendry1981sizes}, we set the last 20\% of entries to -1, representing the inhibitory connections, and the rest to 1 (See Appendix \ref{subsec: dale imp}).\footnote{Our results did not change when we chose a different set of -1 entries instead of the last 20\%.} Unlike \citet{song2016training}, we keep self-connections in the network. Besides biological inspiration, our model also has the following salient features.
% \textcolor{blue}{Unlike \citet{song2016training}, the hidden state of our model is guaranteed to be bounded.}

\begin{figure}[!ht]
\centering
\resizebox{\linewidth}{!}{
\begin{tikzpicture}[
    % GLOBAL CFG
    font=\sf \scriptsize,
    >=LaTeX,
    % Styles
    cell/.style={% For the main box
        rectangle, 
        rounded corners=5mm, 
        draw,
        very thick,
        },
    operator/.style={%For operators like +  and  x
        circle,
        draw,
        inner sep=-0.5pt,
        minimum height =.2cm,
        },
    function/.style={%For functions
        ellipse,
        draw,
        minimum width=0.4cm,
        inner sep=1pt
        },
    ct/.style={% For external inputs and outputs
        circle,
        draw,
        line width = .75pt,
        minimum width=1cm,
        inner sep=1pt,
        },
    gt/.style={% For internal inputs
        rectangle,
        draw,
        minimum width=4mm,
        minimum height=3mm,
        inner sep=1pt
        },
    mylabel/.style={% something new that I have learned
        font=\scriptsize\sffamily
        },
    ArrowC1/.style={% Arrows with rounded corners
        rounded corners=.25cm,
        thick,
        },
    ArrowC2/.style={% Arrows with big rounded corners
        rounded corners=.5cm,
        thick,
        },
    ]
%Start drawing the thing...    
    % Draw the cell: 
    \node [cell, minimum height =3.4cm, minimum width=5.4cm] at (0,0){} ;

   % Draw opérators   named mux# , add# and func#
    \node [function] (tanh1) at (1.5,1.2) {\large $f$};
    \node [function] (alpha) at (-1.2,1.2) {\large $\alpha$};
    \node [function] (minusalpha) at (0,-0.2) {\large $1 - \alpha$};

    % Draw External inputs? named as basis c,h,x
    \node[ct, label={[mylabel]hidden}] (h) at (-4,1.2) {\large $\mathbf{h}^{(t-1)}$};
    \node[ct, label={[mylabel]input}] (x) at (-4,-1.2) {\large $\mathbf{x}^{(t)}$};

    % Draw External outputs? named as basis c2,h2,x2
    \node[ct, label={[mylabel]next\_hidden}] (h2) at (4,1.2) {\large $\mathbf{h}^{(t)}$};

% Start connecting all.
    %Intersections and displacements are used. 
    % Drawing arrows    
    \draw [->, ArrowC2] (h) -- (alpha) -- (tanh1) -- (h2);

    \draw [ArrowC2] (minusalpha) |- (tanh1);
    \draw [ArrowC2] (x) -| (minusalpha);
    \draw [ArrowC2] (h) -| (-2.2,0) |- (-1, -1.2);

    %Outputs

\end{tikzpicture}
}
\caption{Decay RNN cell, comprising of a skip connection and coupled scalar gates.}
\label{fig:drnn}
\end{figure}
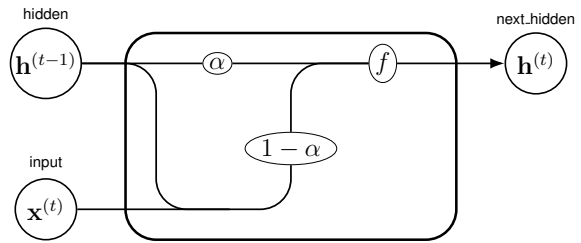

First, the presence of $\alpha$ acts as a coupled gating mechanism to the flow of information (Figure \ref{fig:drnn}), at the same time maintaining an exponential moving average of the hidden state. Thus, $\alpha$ values close to 1 correspond to memories of the distant past. It is worth mentioning that \citet{oliva2017statistical} have considered the exponential moving average in the context of RNNs. However, their approach manually selected a set of scaling parameters, whereas we have a systematic way of arriving at the values of those parameters by making them learnable for the task at hand. 

Second, our model also has an intrinsic skip connection deriving out of its formulation.\citet{yue2018residual} has shown that the architectures with skip connections provide an alternate path for the flow of gradients during the error backpropagation. At the same time presence of coupled gates slows down the vanishing of gradient \cite{bengio2013advances}. Thus, despite of its simple un-gated structure, the features discussed above provide safeguards against vanishing gradient.

To examine the importance of Dale's principle in the learning process, we made a variant of our Decay RNN without Dale's principle, which we call the {\em Slacked Decay RNN} (SDRNN), with updates to \(\mathbf{c}^{(t)}\) made as follows: 
\begin{center}
    $\mathbf{c}^{(t)} = \mathbf{Wh}^{(t-1)} + \mathbf{Ux}^{(t)} + \mathbf{b}$
\end{center}
To understand the role of the correlation between the hidden states in the Decay RNN formulation, we devised an ablated version of our architecture, which we refer to as the {\em Ab-DRNN}. With the following update equation, we remove the mathematical factor (\(\mathbf{Wh}^{(t-1)}\)) that gives rise to a correlation between hidden states:  
\begin{center}
    $\mathbf{h}^{(t)}=f(\alpha\mathbf{h}^{(t-1)} + (1-\alpha)(\mathbf{Ux}^{(t)} + \mathbf{b}))$
\end{center}

\section{Datasets}
\label{subsec: dataset}
For the number prediction (Section \ref{Number Prediction Task}) and grammaticality judgment (Section \ref{subsec: Grammaticality}) tasks, we used a corpus of 1.57 million sentences from Wikipedia \cite{linzen2016assessing}, of which 10\% were used for training, 0.4\% for validation, and the remaining were reserved for testing. On the other hand, for the language modeling task (Section \ref{subsec:Language Modeling}), the model was trained on a 90 million word subset of Wikipedia comprising of 3 million training and  0.3 million validation sentences \cite{gulordava2018colorless}. 

Despite having a large number of training points, these datasets have certain drawbacks, including the lack of a sufficient number of syntactically challenging examples leading to poor generalization over the sentences out of the training data distribution. Therefore, we construct a generalization set as described in \citet{marvin2018targeted}, where we generate the sentences out of templates that can be described using a non-recursive context-free grammar. The use of the generalization set allows us to test on a much broader range of linguistic phenomena. We will use this dataset for the targeted syntactic evaluation of our trained models.

\section{Experiments}
\label{sec:Experiments}
Here we will describe our experiments\footnote{Our code is available at \hyperlink{https://github.com/bhattg/Decay-RNN-ACL-SRW2020}{https://github.com/bhattg/Decay-RNN-ACL-SRW2020}} to assess the models' ability to capture syntax-sensitive dependencies. Details regarding the training settings are available in  Appendix \ref{subsec: training settings}. 

\subsection{Number Prediction Task}
\label{Number Prediction Task}
The number prediction task was proposed by \citet{linzen2016assessing}. In this task, the model is required to predict the grammatical number of the verb when provided a sentence up to the verb.
\begin{enumerate}
    \item The \textbf{path} to success \textbf{is} not straight forward.
    \item The path to success  $\rule{1cm}{0.15mm}$
\end{enumerate}
The model will take the second sentence as input and has to predict the number of the verb (here, singular). Table \ref{Table: Verb Number Prediction} shows the results on the number prediction task. All the models including SRNs performed well on this task. Thus, this indicates that even vanilla RNNs can identify singular and plural words and can associate the main subject with the upcoming verb. 
\begin{table}[!ht]
\centering
\resizebox{0.4\textwidth}{!}{%
\begin{tabular}{l r r}
Model & No. Prediction & Grammaticality\\
\hline
SRN & 97.70& 50.12\\
LSTM &  98.59 & 95.81\\
GRU    &  \textbf{98.81} & 94.26 \\
EIRNN & 94.68 & 84.51 \\
\hline
\textbf{DRNN}  & 98.66 & 95.48 \\
\textbf{SDRNN}  & 98.65&\textbf{96.83} \\
\textbf{Ab-DRNN}  & 97.37& 85.98\\
\end{tabular}}
\caption{\label{Table: Verb Number Prediction}\% Accuracy of models when tested on $\sim$ 1.4 million sentences for the number prediction and grammaticality judgement tasks.}
% Performance of Ab-DRNN suggests the need of exponential moving average in the architectures, which is catered by our Decay RNN.
\end{table}
\subsection{Joint Analysis of Intervening Nouns}
\label{subsec:Joint analysis}
So far in the literature, when looking at intervening material in agreement tasks, the research has tended to focus on agreement attractors, the intervening nouns with the opposite number to the main noun \cite{kuncoro-etal-2018-lstms}. However, we posit that the role of non-attractor intervening nouns may also be important when understanding a model's decisions. For long-range dependencies in agreement tasks, a model may be influenced by the presence of non-attractor intervening nouns instead of purely capturing the verb's relationship with the main subject. Hence an analysis done solely based on the number of agreement attractors may be misleading. 
% Table \ref{tab: joint} shows a comparison between the performance of LSTM and DRNN when compared to the number prediction accuracies.
Table \ref{tab: joint} shows an improvement in the verb number prediction accuracy with an increasing number of non-attractors (n), even as the subject-verb distance and the attractor count are kept fixed. This indicates that the models are also using cues present in the vicinity of the main verb to predict its number, apart from agreement with the main noun.

\begin{table}[!ht]
    \centering
    \resizebox{0.6\linewidth}{!}{%
    \begin{tabular}{l r r r}
         Model& n=0 & n=1 & n=2\\
         \hline
         \textbf{DRNN} & 90.65&95.56&96.06 \\
         LSTM &90.4&95.56&95.63\\
    \end{tabular}}
    \caption{Number prediction \% accuracy with an increasing number of non-attractor intervening nouns (n). The distance between the main subject and the corresponding verb is held constant at 7 and the attractor count at 1.}
    \label{tab: joint}
\end{table}

\subsection{Grammaticality Judgement}
\label{subsec: Grammaticality}
The previous objective was predicting the grammatical number of the verb after providing the model an input sentence only up to the verb. However, this way of training may give the model a cue to the syntactic clause boundaries. In this section, we describe the grammaticality judgment task. Given an input sentence, the model has to predict whether it is grammatical or not. To perform well on this task, the model would presumably need to allocate more resources to determine the locus of ungrammaticality. For example, consider the following pair of sentences\textsuperscript{\ref{note2}} : 
\begin{enumerate}
    \item The \textbf{roses} in the vase by the door \textbf{are} red.
    \item *The \textbf{roses} in the vase by the door \textbf{is} red.
\end{enumerate}
The model has to decide, for input sentences such as the above, whether each one is grammatically correct or not. Table \ref{Table: Verb Number Prediction} shows the performance of different recurrent architectures on this task. It can be seen that SRNs, which were comparable to LSTMs and GRUs on the prediction experiment described in Section \ref{Number Prediction Task}, are no better than random on the grammaticality judgment task. On the other hand, the Ab-DRNN performed better than the SRN. This highlights the importance of a balance between the uncorrelated hidden states (\(\mathbf{h}^{(t)}\)), and the connected hidden states (\(\mathbf{Wh}^{(t)}\)), which is modeled by the Decay RNN. Due to its architectural similarity with the  Independent RNN \cite{li2018independently}, which has independent connections among neurons in a layer, Ab-DRNN did not suffer from the vanishing gradient problem. % At the same time, our balancing effect is analogous to ``squashing" in \citet{mccoy2020does} which explains the better performance of GRU over LSTMs. 

% \begin{minipage}{1.0\textwidth}
%   \strut\newline
%   \centering
%   \includegraphics[scale=0.4]{fullGram.png}
%   \captionof{figure}{caption}\label{fig:figure1}
% \end{minipage}

\subsubsection*{Importance of the generalization set}
\begin{figure}[!ht]
    \centering
    \includegraphics[width=\linewidth, height = 4cm]{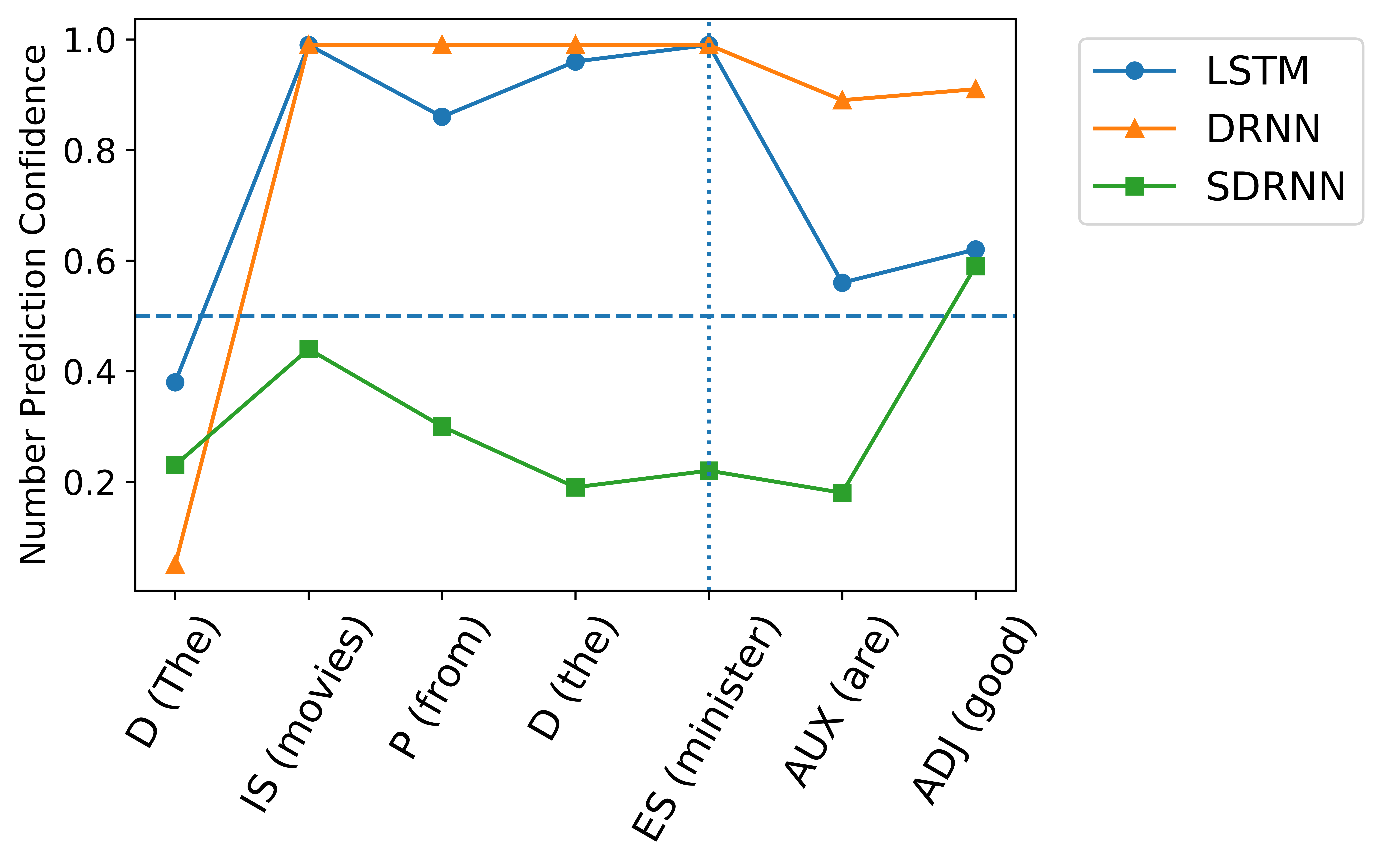}
    \caption{Number prediction confidence (for the correct verb number) averaged over the generalization set (540 sentences) for prepositional phrases with plural inanimate subjects (IS). An example word for each position is indicated in parentheses. Values at ES indicate the confidence for the following verb/auxiliary. For the example sentence, confidence $<0.5$ implies singular verb number prediction, and confidence $>0.5$ plural.}
    \label{fig: pvn confidence}
\end{figure}

\citet{capano2015optimal} had argued that the inclusion of Dale's principle improved generalization abilities for multitask learning. For our models trained on a single task, we use the generalization set to determine the number prediction confidence profile over the sentences. Figure \ref{fig: pvn confidence} describes the average number prediction confidence at each part of speech for all prepositional phrases with inanimate subjects. We note the anomalously low confidence of the SDRNN at plural inanimate subjects (like `movies', `books'), unlike the DRNN.

\begin{table*}[ht]
    \centering
    \resizebox{0.72\textwidth}{!}{%
    \begin{tabular}{l r r r r r r r}
         & SRN & GRU & LSTM &\textbf{DRNN}& \textbf{SDRNN} &\textbf{Ab-DRNN}& ON-LSTM \\
         \hline
         \hline
         Validation Perplexity & 114.74& 53.78& 52.73&76.67&76.88&86.42&- \\
         Parameters &  1.4M &  4.2M &  5.6M &  1.4M&  1.4M & 0.55M &- \\
         \hline
         \hline
         \textbf{Short-Range Dependency} \\
         \hline 
         \textbf{SV Agreement:} \\
         Simple &0.88 &0.95 &0.92 &0.95 &{0.97} & 0.90& \textbf{0.99}\\
         Sentential Complement &0.84 &0.86 &{0.93 }&0.89 &0.92& 0.85&\textbf{0.95}\\
         Short VP Coord &0.5 &{0.87} &0.85 &0.73 &0.77 &0.69& \textbf{ 0.89}\\
         In an object RC & 0.59&0.75 &\textbf{0.87 }&0.77 &0.74 & 0.63&0.84\\
         In an object RC (no that) &0.57&0.67&{0.75}&0.74&0.71 &0.62& \textbf{ 0.78}\\
        \hline
        \textbf{Reflexive Anaphora: }\\
        Simple &0.51&{0.85}&{0.85}&0.75&0.73& 0.63& \textbf{0.89}\\
        Sentential Complement &0.56&0.78&{0.83}&0.68&0.65&0.62&\textbf{0.86 }\\
        \hline
        \textbf{Negative Polarity Items :} \\
        Simple (grammatical vs. intrusive) &0.01&0.51&\textbf{0.56}&0.25&0.01&0.29&0.18\\
        Simple (intrusive vs. ungrammatical) &\textbf{0.7}&0.66&0.48&0.54&0.5& 0.51 &0.5\\
        Simple (grammatical vs. ungrammatical) & 0.11&\textbf{0.67}&0.55&0.45&0.38&0.31&0.07 \\
        \hline 
        \hline 
        \textbf{Long-Range Dependency} \\
        \hline 
        \textbf{SV Agreement:} \\
        Long VP coordination &0.51&\textbf{0.8}&\textbf{0.8}&0.55&0.62& 0.51& 0.74\\
        Across a PP &0.51&\textbf{0.75}&0.6&0.56&0.54& 0.53& 0.67\\
        Across a subject RC &0.52&\textbf{0.67}&\textbf{0.67}&0.53&0.55&0.52&
        0.66 \\
        Across an object RC  & 0.51& 0.51& 0.55&\textbf{0.64}&0.58& 0.57&0.57\\
        Across an object RC (no that) &0.50&0.50&0.51&\textbf{0.65}&0.60& 0.59& 0.54\\
        \hline
        \textbf{Reflexive Anaphora :} \\
        Across a RC &0.51&0.58&0.57&0.62&\textbf{0.66}& 0.58& 0.57\\
        \hline
        \textbf{Negative Polarity Items:}\\
        Across a RC (grammatical vs. intrusive) &\textbf{0.87}&0.55&0.55&0.32&0.48&0.57&0.59 \\
        Across a RC (intrusive vs. ungrammatical) &0.02&0.29&0.22&\textbf{0.5}&0.37&0.36&0.20\\
        Across a RC (grammatical vs. ungrammatical) &0.1&0.2&0.03&0.1&\textbf{0.3}&0.11&0.11 \\
        \hline 
        \hline
        Mean Arithmetic Rank &5.94&3&3.31&3.52&3.68&4.73&\textbf{2.94}\\
    \end{tabular}}    \renewcommand\thetable{4}
    \caption{Accuracy of models on targeted syntactic evaluation. RC: Relative Clause, PP: Prepositional Phrase, VP : Verb Phrase. Closeness in the mean arithmetic rank of models (other than SRNs) across tasks suggests that within the current space of sequential recurrent models, none dominates the others.}
    \label{table: Targeted syntactic evaluation}
\end{table*}

\begin{table}[H]
\centering
\resizebox{0.8\linewidth}{!}{%
\begin{tabular}{l r r}
Task & DRNN & SDRNN\\
\hline
Across object RC (no that) anim & \textbf{0.45}&0.28 \\
Reflexive Sentential Comp. &\textbf{0.65}&0.6 \\
Long VP Coordination &\textbf{0.53}&0.43
\end{tabular}}
\renewcommand\thetable{3}
\caption{\label{Table: syn on gram} Accuracy comparison of DRNN and SDRNN when tested on the generalization set for the grammaticality judgement task; `anim' refers to an animated noun.}
\end{table}

In Table \ref{Table: syn on gram},\footnote{Here, we present three tests from the targeted syntactic evaluation framework. Others test results can be found in Appendix \ref{subsec: app_gram}.} we present the result of the models trained for the grammaticality judgment task and tested on the synthetic generalization set. From the results, we can see that despite having nearly the same accuracy on the original testing data (Table \ref{Table: Verb Number Prediction}), there is a substantial difference in the generalization accuracies of the DRNN and SDRNN. The DRNN shows better generalization than the SDRNN in the experiments mentioned in Table \ref{Table: syn on gram} and Figure \ref{fig: pvn confidence}. This might be due to regularising effects induced by Dale's constraint. This is an interesting observation that merits further investigation.  

\subsection{Language Modeling}
\label{subsec:Language Modeling}
Word-level language modeling is a task that helps in the evaluation of the model's capacity to capture the general properties of language beyond what is tested in specialized tasks focused on, e.g., subject-verb agreement. We use perplexity to compare our model's performance against standard sequential recurrent architectures. Table \ref{table: Targeted syntactic evaluation} shows the validation perplexity of different language models along with the number of learnable parameters for the task. From the Table \ref{table: Targeted syntactic evaluation}, we observe that incorporating the components of the Ab-DRNN and the SRN in a coupled way might have led to the improved performance of the Decay RNN.
% Decay RNN and SRN differ by just a parameter.
% The slight modification in the architecture of SRN to get DecayRNN, led to the exceedingly well improvement in the validation perplexity of the model. The improved generalization abilities of Ab-DRNN as shown in Table \ref{table: Targeted syntactic evaluation}, highlights the need of having free 

\subsection{Targeted Syntactic Evaluation}
\label{subsec: Targeted Syntactic Evaluation}
Targeted syntactic evaluation \cite{marvin2018targeted} is a way to evaluate the language model across different classes of structure-sensitive phenomena. This includes subject-verb agreement, reflexive anaphora, and negative polarity items (NPI).\footnote{The definitions of these linguistic terms are provided in the supplementary material of \citet{marvin2018targeted}.}
Table \ref{table: Targeted syntactic evaluation} shows that even with a simple architecture, the Decay RNN class of models performs fairly similarly to LSTMs and much better than SRNs for many tests.\footnote{Results for the ON-LSTM are directly quoted from \citet{shen2018ordered}.} In the case of long-range dependencies and NPI involving relative-object clauses, our models perform substantially better than LSTMs. High variability in the performance of the models in the case of NPIs might be due to non-syntactic cues as pointed out by \citet{marvin2018targeted}. Based on the mean ranks observed in Table \ref{table: Targeted syntactic evaluation}, we conjecture that there is no sequential recurrent structure at present which outperforms the others across the board. However, SRNs alone are not sufficient for most purposes.  

\section{Conclusion}
In this paper, we proposed the Decay RNN, a bio-inspired recurrent network that emulates the decaying nature of neuronal activations after receiving excitatory and inhibitory impulses from upstream neurons. We have found that the balance between the free term (\(\mathbf{h}^{(t)}\)) and the coupled term (\(\mathbf{Wh}^{(t)}\)) enabled the model to capture syntax-level dependencies. As shown by \citet{mccoy2020does,kuncoro-etal-2018-lstms}, explicitly modeling hierarchical structure helps to discover non-local structural dependencies. The contrast in the performance of the language models encourages us to look at the inductive biases, which might have led to better syntactic generalization in certain cases.
Recently, \citet{maheswaranathan2020recurrent} showed the existence of a line attractor in the dynamics of the hidden states for sentiment classification. Thus, similar dynamical-system-based analysis can be extended to our settings to further understand the working of the Decay RNN.
%From the optimization point of view, it is yet not possible to incorporate Dale's principle for these artificial neural networks \cite{Lillicrap2020}. Our work does provide a partial way to solve this problem; however, better ways to incorporate Dale's constraint with back-propagation is one of our future works.
%One way is to investigate the locus of error in case of a number prediction task, by analyzing the patterns in the decision of the models at each instance.

From the cognitive neuroscience perspective, it would be interesting to investigate if the proposed Decay RNN can capture some aspects of actual neuronal behaviour and language cognition. Our results here do at least indicate that the complex gating mechanisms of LSTMs (whose cognitive plausibility has not been established) may not be essential to their performance on many linguistic tasks, and that simpler and perhaps more cognitively plausible RNN architectures are worth exploring further as psycholinguistic models.

\section*{Acknowledgements}
We wish to thank the anonymous reviewers, and Jakob Prange and ACL SRW for the post-acceptance mentorship program; Pankaj Malhotra for valuable comments on earlier versions of this paper; and Tal Linzen for helpful discussion. 
\bibliography{acl2020}
\bibliographystyle{acl_natbib}

\clearpage

\appendix
\section{Appendix}

\subsection{Effect of agreement attractors}
\begin{figure*}
  \centering
  \includegraphics[width = \linewidth]{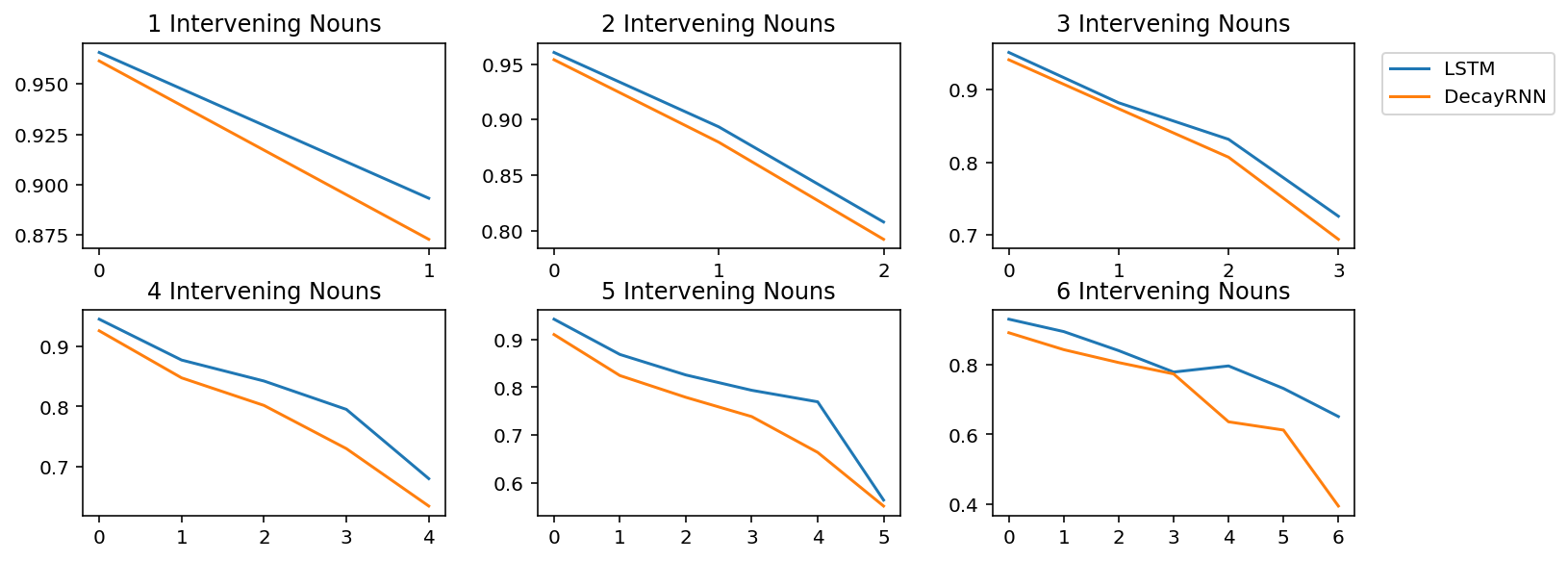}
  \caption{Trends in the performance of the LSTM (blue) and DRNN (orange) models with increasing numbers of intervening nouns. For each subplot corresponding to a fixed intervening noun number, the number of agreement attractors increases as we move from left to right on the $x$-axis.}\label{fig:figure1}
\end{figure*}

In this section, we present the trends in the testing performance of the LSTM and the Decay RNN (DRNN) for the grammaticality judgment task. Figure \ref{fig:figure1} shows the performance of the models when we fix the number of intervening nouns and vary the count of attractors between the main subject and the corresponding verb. The decreasing performance of the models with the introduction of more attractors indicates that they cause the models to get more confused about the upcoming verb number.

% \begin{figure}[b]
%     \centering
%     \includegraphics[scale=0.4]{fullGram.png}
%     \caption{caption}
%     \label{fig: figure1}
% \end{figure}

% \subsection{Description of POS tags in generalization set}

% The generalization set tests the ability of our models to perform well on the sentences, which are not sampled from the training data distribution. 
% \begin{table}[!h]
%     \centering
%     \resizebox{1\linewidth}{!}{%
%     \begin{tabular}{c|c|c}
%          Part of Speech& Description & Example\\
%          \hline
%          MS & Animate main subject & authors, pilots, surgeons  \\
%          ES & Animate intervening subject & guards, chefs\\
%          EV & Intervening verbs & like, hate\\  
%          IS & Inanimate subject & movies, books\\
%          P, IP & Preposition & to, behind, on\\ 
%          D & Determiner & a, an, the\\
%          MV & Main verb & is tall, is going\\
%          LMV & Long verb phrase(for long vp coordination)& knows many different foreign languages
%     \end{tabular}}
%     \caption{}
%     \label{tab: joint}
% \end{table}

\subsection{Comparison between DRNN and SDRNN}
\label{subsec: app_gram}
\begin{figure*}
    \centering
    \includegraphics[width = \linewidth]{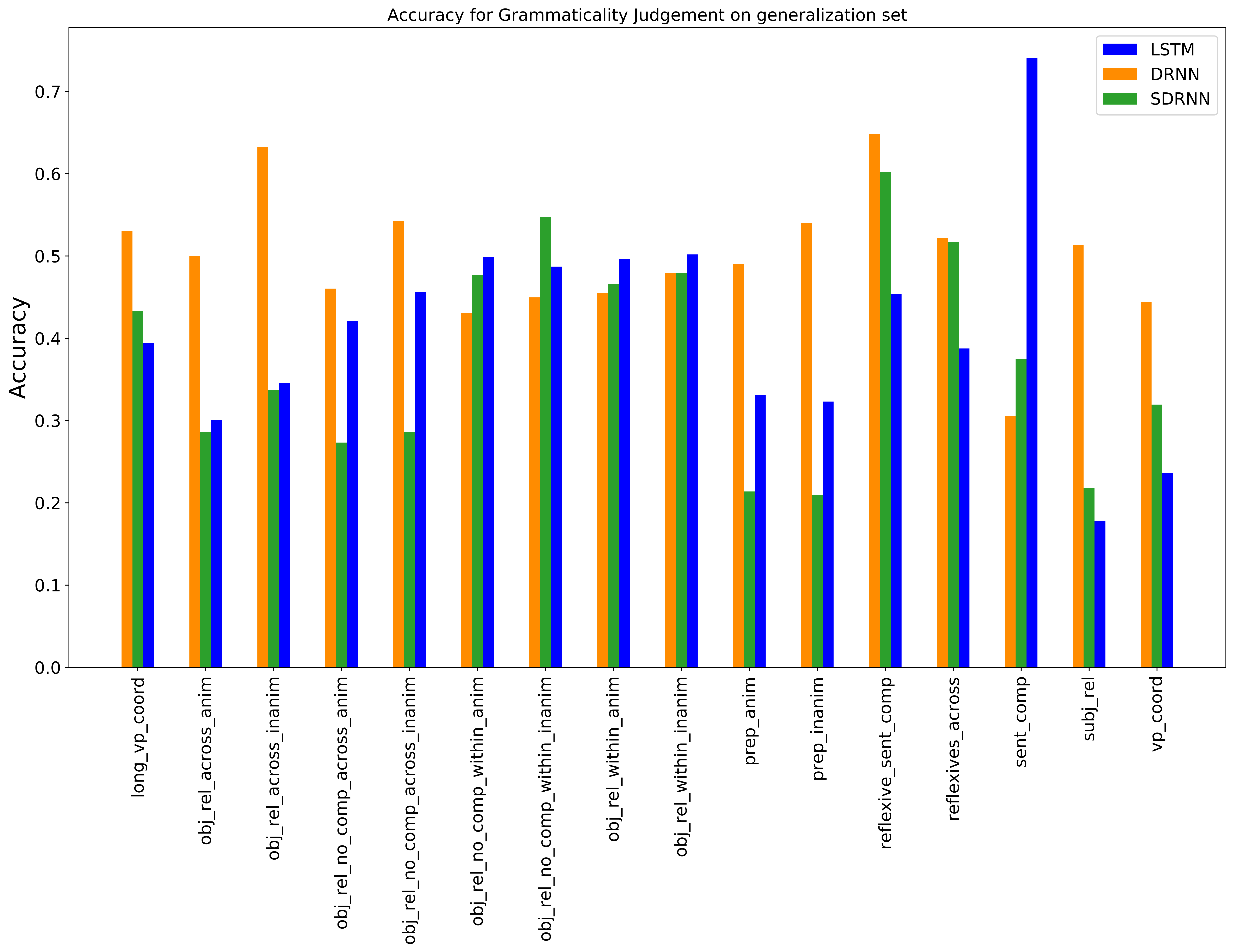}
    \caption{Performance of the LSTM (blue), DRNN (orange), and SDRNN (green) models for the different types of sentences in the generalization set, when trained for the grammaticality judgment task. There were at least 200 test sentences for each of these types.}
    \label{figure2}
\end{figure*}

In Section \ref{subsec: Grammaticality}, we saw that in terms of testing accuracy for grammaticality judgment, the Slacked Decay RNN (SDRNN) outperformed the Decay RNN (DRNN). For a robust investigation of this behaviour, we tested our models on the generalization set and mentioned a subset of our results on grammaticality judgment in Table \ref{Table: syn on gram}. Here we present a bar graph (Figure \ref{figure2}) depicting the model performance when tested on the generalization set for the grammaticality judgment task. A substantial difference in the performance of the SDRNN and the DRNN reinforces the possibility of the regularizing effects of Dale's principle.

\subsection{Implementation of Dale's constraint}
\label{subsec: dale imp}

\[\forall w_{i,j} \in ReLU(\textbf{W}), w_{i,j} \geq 0
\]
\[\textbf{ReLU(W)W}_{dale} = 
\left[
\begin{array}{cccc}
w_{1,1} & w_{1,2} & \dots & w_{1,n}\\
w_{2,1} & w_{2,2} & \dots & w_{2,n}\\
\vdots & \vdots & \vdots & \vdots\\
w_{n,1} & w_{n,2} & \dots & w_{n,n}\\
\end{array} \right ]
\left[
\begin{array}{cccc}
1 & 0 & \dots  & 0\\
0 & 1 & \dots & 0\\
\vdots & \vdots & \vdots & \vdots\\
0 & 0 & \dots & -1\\
\end{array}\right] = 
\left[
\begin{array}{cccc}
+ & + & \dots & - \\
+ & + & \dots & - \\
\vdots & \vdots &  \vdots\\
+ & + & \dots & - \\
\end{array} \right]
\]

\subsection{Training settings}
\label{subsec: training settings}
For the number prediction task and the grammaticality judgment task the network is trained as a binary classifier. The network is single-layered, with ReLU activation and trained with embedding and hidden layer dimension being 50, and a batch size of 1. We have reported the average accuracies after 3 separate runs in Table \ref{Table: Verb Number Prediction}. For targeted syntactic evaluation, we have trained a language model to predict the grammaticality of a sentence.  In our language model, we used a 2-layered network with $tanh$ activation, a dropout rate of 0.2 with embedding dimension 200, hidden dimension 650, and a batch size of 128. All models are trained with a learning rate of 0.001 using the Adam optimizer \cite{kingma2014adam}. 

\subsection{Decay parameter ($\alpha$) learning}

\begin{figure*}
    \centering
    \includegraphics{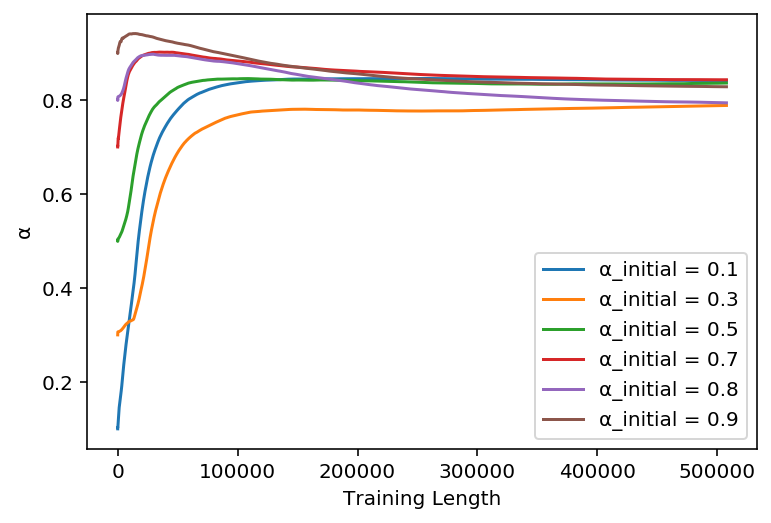}
    \caption{Moving average of $\alpha$ over the course of training for different initializations. 1 unit of training length is 1 forward pass.}
    \label{figure3}
\end{figure*}

In the main text, we describe the balancing effect of $\alpha$ in the Decay RNN model. We present the trend in the learned value of $\alpha$ throughout training for the grammaticality task for various initializations in Figure \ref{figure3}. We observe that for all $\alpha$ initializations in the range (0,1), the learned value converges to around 0.8. Hence, we initialize our $\alpha$ to 0.8 at the start of the training process. 

\end{document}